\documentclass[conference]{IEEEtran}
\IEEEoverridecommandlockouts
\usepackage{cite}
\usepackage{amsmath,amssymb,amsfonts}
\usepackage{graphicx}
\usepackage{textcomp}
\usepackage{xcolor}
\usepackage{bm}
\usepackage{epsfig}
\usepackage{amsmath}
\usepackage{amssymb}
\usepackage{color}
\usepackage{url}
\usepackage{multirow}
\newlength{\figurewidth}
\newlength{\smallfigurewidth}
\newcommand{\eatme}[1]{ }
\usepackage{paralist}
\usepackage{adjustbox}
\usepackage{wrapfig}
\usepackage{tikz}

\usepackage{cite}
\usepackage{amsmath,amssymb,amsfonts}
\usepackage{algorithm}
\usepackage{graphicx}
\usepackage{textcomp}
\usepackage{xcolor}
\usepackage{subcaption}
\usepackage{siunitx}

\usepackage[T1]{fontenc}
\usepackage{graphicx}
\usepackage{booktabs}
\usepackage{longtable}
\usepackage{pbox}
\usepackage{mathtools}
\usepackage{caption}
\usepackage{subcaption}
\usepackage{algpseudocode}
\usepackage{algorithm}
\usepackage{siunitx}
\usepackage[inline]{enumitem}
\usepackage{siunitx}
\usepackage{diagbox}
\usepackage[]{mdframed}
\usepackage{comment}
\usepackage[normalem]{ulem}

\usepackage{xparse}
\NewDocumentCommand\N{sm}{\mathcal{N}\IfBooleanT#1{^{\ast}}_{#2}}

\makeatletter
\NewDocumentCommand{\@Coefficients}{m}{\text{\ttfamily\upshape #1}}
\newcommand\uMultilevelCoefficients{\@Coefficients{u\char`_mc}}
\newcommand\newMultilevelCoefficients{\@Coefficients{\~{u}\char`_mc}}
\makeatother

\begin{document}

\title
{{Attention Based Machine Learning Methods for Data Reduction with Guaranteed Error Bounds\thanks{This work was partially supported by DOE RAPIDS2 DE-SC0021320 and DOE DE-SC0022265.}}
} 

\author{
Xiao Li, Jaemoon Lee, Anand Rangarajan, and Sanjay Ranka  \\
\begin{minipage}{\linewidth}
    \begin{center}
        \begin{tabular}{c}
            \textit{Department of Computer and Information Science and Engineering }\\
            \textit{University of Florida, Gainesville, FL 32611}\\
            %\text{\{xiao.li, j.lee1\}@ufl.edu}, \text{\{anand, ranka\}@cise.ufl.edu}
        \end{tabular}
    \end{center}
\end{minipage}
}

\maketitle

\thispagestyle{empty}
\begin{abstract}
    
Scientific applications in fields such as high energy physics, computational fluid dynamics,  and climate science  generate vast amounts of data at high velocities. This exponential growth in data production is surpassing the advancements in computing power, network capabilities, and storage capacities. To address this challenge, data compression or reduction techniques are crucial. These scientific datasets have underlying data structures that consist of structured and block structured multidimensional meshes where each grid point corresponds to a tensor. It is important that data reduction techniques leverage strong spatial and temporal correlations that are  ubiquitous in these applications. Additionally, applications such as CFD, process tensors comprising  hundred plus species and their attributes at each grid point. Reduction techniques should be able to leverage interrelationships between the elements in each tensor. 

In this paper, we propose an attention-based hierarchical compression method utilizing a block-wise compression setup. We introduce an attention-based hyper-block autoencoder to capture inter-block correlations, followed by a block-wise encoder to capture block-specific information. A PCA-based post-processing step is employed to guarantee error bounds for each data block. Our method effectively captures both spatiotemporal and inter-variable correlations within and between data blocks. Compared to the state-of-the-art SZ3, our method achieves up to 8$\times$ higher compression ratio on the multi-variable S3D dataset. When evaluated on single-variable setups using the E3SM and XGC datasets, our method still achieves up to 3$\times$ and 2$\times$ higher compression ratio, respectively.

\end{abstract}

\section{INTRODUCTION}

Scientific simulations often generate vast amounts of data, typically several terabytes per run, while tracking hundreds of variables. For instance, S3D \cite{Yoo11}, a direct numerical simulation code for turbulent combustion, produces detailed datasets that include species concentrations, temperature, pressure, and velocity fields. Storing and managing such high-dimensional, large-size data pose significant challenges. In response to these challenges, scientific data compression has emerged as a crucial area of research over the past decade \cite{cappello2019use, gong2023mgard, SZ3}. Concomitantly, neural network architectures for data reduction and compression have seen considerable evolution as well within the same time frame \cite{liu2023srn}. Consequently, we have seen numerous architectures for scientific data compression. This work is squarely in the transform architecture camp and in particular marries self-attention with error-bound guarantees (on the reconstructions).  Previous data reduction and compression methods using neural networks have largely ignored the guaranteed error bounds aspect central to this work \cite{liu2023learned, guo2023compression}. This requires processing reconstruction residuals such that they satisfy error bounds. 

Scientific datasets often feature underlying data structures composed of structured and block-structured multidimensional meshes, where each grid point corresponds to a tensor. Effective data reduction techniques need to make use of the strong spatial and temporal correlations that are common in such applications. In fields like computational fluid dynamics (CFD), these datasets include tensors at each grid point representing numerous species and their attributes. Therefore, reduction techniques should also leverage the interrelationships between the elements within each tensor to enhance efficiency and accuracy.

%This process dovetails with the choice of compression architectures since there is no clear winner in this category. After much experimentation with different architectures, we empirically discovered that a coarse-to-fine tensor blocks architecture utilizing self-attention gave the best performance. This architecture is now described.

To leverage the above observations, the fundamental point of departure in the present work relative to previous work is the coarse-to-fine decomposition of the set into hyper-blocks, blocks and vectors (with the last subjected to the error bounding process described above). Since the cardinality of the hyper-block set is less than the rest, we elected to apply self-attention based autoencoders for an initial stage of data reduction. That is, we construct encoders and decoders with self-attention layers which operate on hyper-blocks. Subsequently, we process the residuals from the hyper-block stage block by block using a second autoencoder which eschews attention (since this is not really required for residual processing). Finally, at the individual vector level, 
% (on the S3D application) and build a very simple fully connected architecture which is capable of bounding the reconstruction error of each vector. The other two applications do not feature the full breadth of the coarse-to-fine architecture, and here 
we use principal component analysis (PCA) for the error bounding process. The contributions of this paper can be summarized as follows:
\begin{itemize}
    \item We propose a hierarchical approach that leverages both short-range and long-range correlations in spatiotemporal grid of tensors as well as correlations within a tensor. This is achieved by first building an attention based architecture to capture long-range correlations across multiple temporal blocks (also called a hyper-block) and then performing a spatiotemporal blocking of tensors to capture short-range correlations. This novel architecture results in significantly higher compression than existing methods for a wide range of scientific datasets.

    \item We introduce an attention-based encoder decoder architecture that encodes each block and then uses attention between multiple blocks of a hyper-block. The self-attention mechanism dynamically weighs the relationships between blocks, enabling the model to capture long-range dependencies and fine-grained correlations within the hyperblock, resulting in more efficient compression. To the best of our knowledge, this is the first application of an attention-based autoencoder for scientific data compression.
    
    \item We propose a PCA-based method to guarantee error bounds at the block level (rather than at the hyper-block level). By retaining only the minimal number of PCA coefficients, our method ensures that each data block meets the user-defined error tolerance while still achieving a competitive compression ratio.
    % This results in significantly low computational overhead as the size of the basis vector is limited to each block (of a smaller size than the hyper-block).

\end{itemize}
Overall, our compression ratios for a given NRMSE error are significantly better than those of existing methods. For instance, compared to the state-of-the-art compressor SZ3 \cite{SZ3}, our method achieves up to 8$\times$ higher compression on the multi-variable S3D \cite{Yoo11} dataset. Additionally, we achieve up to 3$\times$ and 2$\times$ higher compression ratios on the E3SM \cite{e3sm} and XGC \cite{chang2008spontaneous} datasets using single variable, respectively. It is larger for datasets with higher dimensional tensors.
% For a given compression ratio our NRMSE errors could be 2 to 8 times better depending on the dataset. 

\section{Methodology}\label{sec:problem}

%\subsection{Error Bound Guarantee}

\eatme{
\subsection{Challenges in Scientific Data Compression}

Scientific data compression presents a unique set of challenges due to the inherent complexity and characteristics of the data. These challenges are distinct from those encountered in more conventional data types. Unlike videos and images, which typically have two or three dimensions (spatial and temporal), scientific data often spans multiple dimensions, such as space, time, and various physical variables. This high dimensionality makes it challenging to apply standard compression techniques, as these methods are often optimized for lower-dimensional data. Capturing and compressing the correlations across many dimensions requires more sophisticated models that can handle the complexity of the data structure.
}
%\subsection{A Solution for High-Dimensional Data Compression}

Given that scientific data often exhibits strong local correlations, a natural approach to compress high-dimensional data is to divide the original dataset into multi-dimensional blocks and compress each block separately. However, this approach presents a trade-off: While larger blocks can capture broader correlations, they also introduce significant computational and memory overhead. The increased complexity associated with large blocks can lead to instability during training, slower convergence, and challenges in model tuning. On the other hand, very small blocks may not capture sufficient contextual information. This can result in inefficient compression.
To address these challenges, we propose a hierarchical machine learning technique for scientific data compression. Our approach employs a block-based compression strategy, dividing the original data into appropriately sized multi-dimensional blocks. We integrate an attention mechanism to capture and leverage inter-block correlations. Following this, a block-wise autoencoder is used to effectively capture block-specific information including local and multi-dimensional correlations within each block. This combination aims to optimize both the compression ratio and computational complexity. 

Considering a data point or data block \( x_i \in \Omega \) and the corresponding reconstructed data \( x_i^R \in \Omega^R \), given the user-defined error bound \( \tau \), the goal of ensuring this error bound is to minimize the difference between \( x_i \) and \( x_i^R \) such that the error introduced by compression does not exceed \( \tau \). Specifically, the objective can be formulated as:
\begin{equation}
    \|x_i - x_i^R\| \leq \tau \quad \forall i
\end{equation}
where \( \|\cdot\| \) denotes the appropriate norm (e.g., absolute value for scalars or \( \ell_2 \)-norm for vectors/tensors). This condition ensures that the reconstructed data \( \Omega^R \) remains within a tolerable range of the original data \( \Omega \), preserving the accuracy and reliability of the data for subsequent analysis.

\begin{figure*}[h!]
    \centering
    \includegraphics[width=\textwidth]{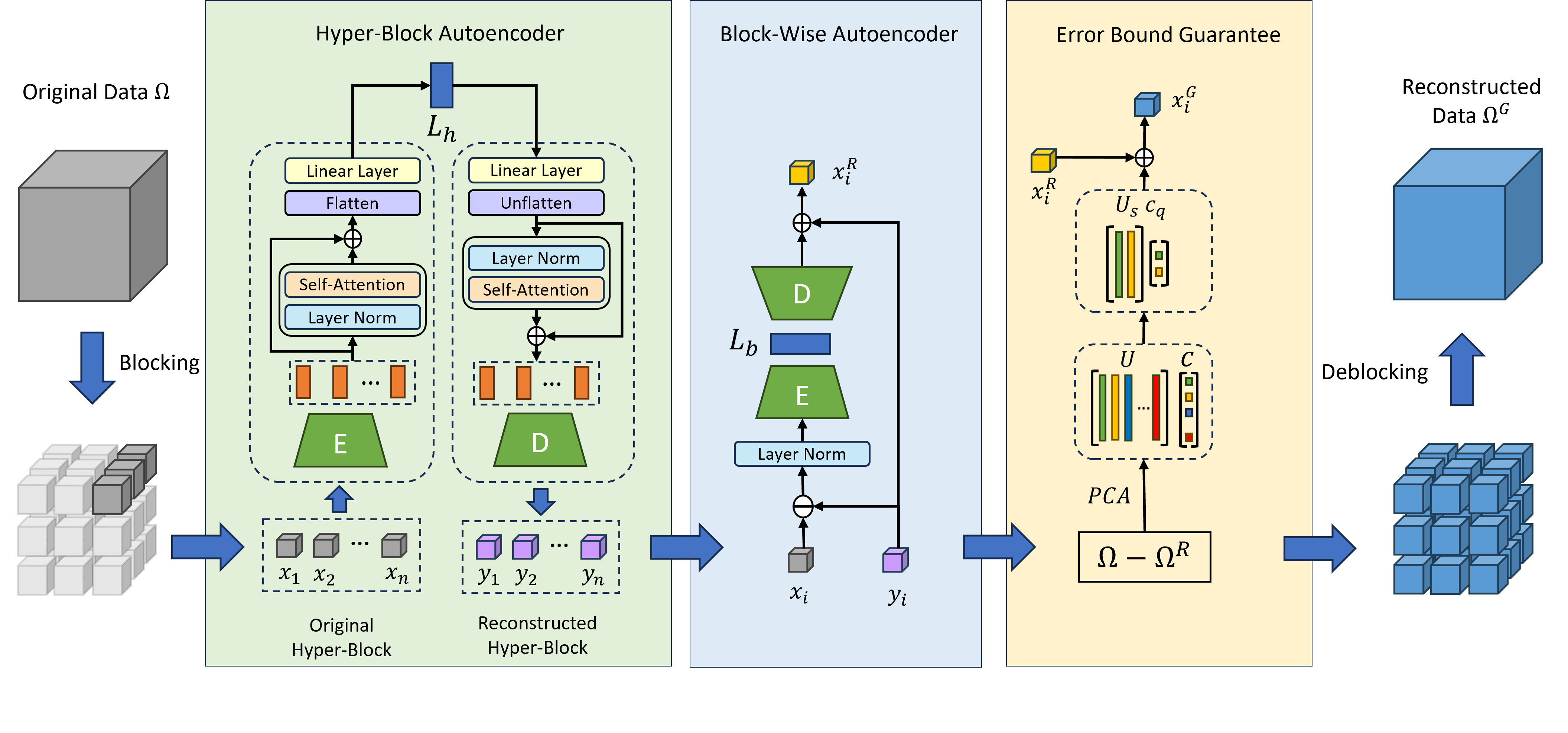}  % Replace with the actual path to your image
    \vspace*{-1.2cm}
    \caption{Method Overview: The original data is divided into hyper-blocks. An attention-based autoencoder is employed to capture inter-block correlations within these hyper-blocks. The residual data from the hyper-block Autoencoder is then processed block by block using a block-wise autoencoder for finer reconstruction. Finally, the reconstructed data undergoes a PCA-based method to ensure the error of each data patch is within guaranteed bounds.}
        \vspace*{-0.5cm}
    \label{fig:pipeline}
\end{figure*}

%\section{Methodology}\label{sec:methodology}
The overview of the proposed method is illustrated in Figure \ref{fig:pipeline}. The compression pipeline can be divided into three components: the hyper-block autoencoder (HBAE), block-wise autoencoder (BAE), and error bound guarantee. First, the original data is divided into blocks, which are then grouped into hyper-blocks based on neighborhood information. These hyper-blocks are fed into the hyper-block autoencoder for coarse compression. Next, the residuals between the original hyper-blocks and the reconstructed hyper-blocks are calculated. This residual data is then processed block by block using the block-wise autoencoder, resulting in finer reconstruction. Finally, we use a PCA-based method to ensure the error of each block remains within acceptable bounds. In the following sections, we will introduce each of these components in detail.

\subsection{Self-Attention Mechanism}

\begin{figure}[h!]
    \centering
    \includegraphics[width=0.4\textwidth]{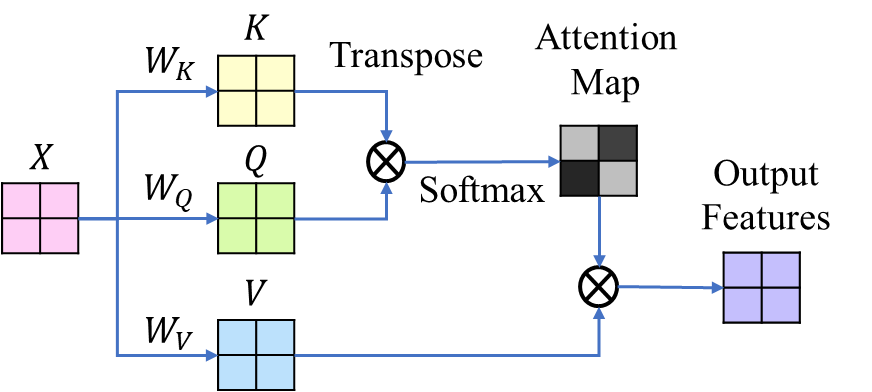}  % Replace with the actual path to your image
    \caption{Self-Attention Mechanism}
    \label{fig:selfatten}
    \vspace*{-0.5cm}
\end{figure}

The self-attention mechanism allows a model to weigh and aggregate information from different parts of a sequence, effectively capturing the contextual information within the data. Given an input sequence represented by an embedding matrix $\boldsymbol{X} \in \mathbb{R}^{n \times d}$, where $n$ is the length of the sequence and $d$ is the dimension of the embedding. The self-attention mechanism first maps an intermediate embedding into three embeddings: query $\boldsymbol{Q} \in \mathbb{R}^{n \times d_k}$, key $\boldsymbol{K} \in \mathbb{R}^{n \times d_k}$ and value $\boldsymbol{V} \in \mathbb{R}^{n \times d_v}$ through linear transformations as follows:
\begin{equation}
    \boldsymbol{Q} = \boldsymbol{X} \boldsymbol{W}_Q, \quad \boldsymbol{K} = \boldsymbol{XW}_K, \quad \boldsymbol{V} = \boldsymbol{XW}_V, \\
\end{equation}
where $\boldsymbol{W}_Q$, $\boldsymbol{W}_K$ and $\boldsymbol{W}_V$ are learned weight matrices with dimensions $\mathbb{R}^{d \times d_k}$, $\mathbb{R}^{d \times d_k}$, and $\mathbb{R}^{d \times d_v}$, respectively. 

The attention scores are computed by taking the dot product of the query matrix $\boldsymbol{Q}$ with the transposed key matrix $\boldsymbol{K}^T$, followed by scaling by dividing ${\sqrt{d_k}}$ to stabilize the scores. A softmax function is then applied to normalize the scores so they sum to 1. The value vectors $\boldsymbol{V}$ are weighted by these attention scores, determining which information should be aggregated from different parts of the sequence. The weighted sum of these value vectors gives the final output of the self-attention mechanism:
\begin{equation}
    Attention(\boldsymbol{Q},\boldsymbol{K},\boldsymbol{V}) = Softmax(\frac{\boldsymbol{QK}^{T}}{\sqrt{d_k}})\boldsymbol{V}
\end{equation}
Compared to simpler models like RNNs or CNNs, the self-attention mechanism introduces higher computational complexity, scaling quadratically with the sequence length. Specifically, computing the dot product between the query and key matrices has a computational complexity of $\mathcal{O}(n^2 \cdot d_k)$, and multiplying the resulting attention scores by the value matrix adds a complexity of $\mathcal{O}(n^2 \cdot d_v)$, leading to an overall complexity of: 
\begin{equation} 
\mathcal{O}(n^2 \cdot d_k + n^2 \cdot d_v). 
\end{equation} 
This makes the self-attention mechanism more expensive for long sequences compared to RNNs or CNNs, which have a linear complexity of $\mathcal{O}(n \cdot d)$. However, the model size cost grows only linearly with the embedding dimension $d$. The total number of learned parameters for the weight matrices $\boldsymbol{W}_Q$, $\boldsymbol{W}_K$, and $\boldsymbol{W}_V$ is: \begin{equation} \mathcal{O}(d \cdot (2d_k + d_v)). \end{equation} While self-attention increases computational cost, it maintains a relatively compact model size, offering an effective trade-off for capturing long-range dependencies without significantly expanding the model. This balance makes it particularly suited for data compression.

\subsection{Hierarchical Models for Compression}

\subsubsection{Hyper-Block Autoencoder}

Previous work in \cite{jaemoon} addresses block-by-block compression. However, a notable drawback of this method is its inability to capture correlations between blocks, leading to suboptimal compression performance. An alternative approach is to compress each block individually and then use an additional layer to capture inter-block dependencies. In this work, we develop an attention-based autoencoder to capture these inter-block correlations. We denote the original data as \(  \Omega  = \{\bm{x}_1, \bm{x}_2, \dots, \bm{x}_N\} \), where \(  \Omega \) consists of \( N \) data blocks, and each \( \bm{x}_i \) represents a single block. Given an encoder \( E(\cdot) \) that projects a high-dimensional data block into an embedding vector, and a decoder \( D(\cdot) \) that reconstructs a data block from an embedding vector. Specifically, we use two fully connected layers with ReLU activation in the middle as the embedding encoder and decoder. The encoder projects the data from its original input dimension to a 128-dimensional embedding, while the decoder takes the 128-dimensional embedding and projects it back to the original dimension. To encode these data blocks, we first group \( k \) data blocks into one hyper block, typically along the temporal dimension, denoted as \( \bm{x}_{i:i+k} \). Each block in the hyper-block \( \bm{x}_{i:i+k} \) is independently fed into the encoder \( E \), resulting in \( k \) embedding vectors, denoted as \( \{ \bm{e}_{i}, \bm{e}_{i+1}, \dots, \bm{e}_{i+k-1} \} \) or collectively as \( \bm{e}_{i:i+k} \). These embeddings capture the essential features of each block within the hyper-block.

To further enhance these embeddings, we apply Layer Normalization \cite{ba2016layer} to \( \bm{e}_{i:i+k} \), which helps stabilize and normalize the values across the embeddings, ensuring that they are on a comparable scale. The normalized embeddings are then fed into a self-attention module, which allows the model to capture dependencies and relationships across different blocks within the hyper-block, resulting in the attention-enhanced embedding vectors, denoted as \( \bm{\tilde{e}}_{i:i+k} \). To preserve the original information from the embeddings, a residual connection is added, directly connecting the input block embeddings \( \bm{e}_{i:i+k} \) to the output of the self-attention module. Finally, each attention-enhanced embedding vector \( \bm{\tilde{e}}_i \) can be expressed as:
\begin{equation}
    \label{eq:atten}
    \bm{\tilde{e}}_i = {Atten}({norm}(\bm{e}_{i:i+k}) )_i + \bm{e}_i,
\end{equation}
where `\( \text{norm} \)' denotes Layer Normalization and `\( \text{Atten} \)' denotes the self-attention mechanism. Although we obtain high-quality embeddings of the original blocks, it is not storage efficient to save these embeddings for data reconstruction. Therefore, we flatten \( \bm{\tilde{e}}_{i:i+k} \) into a vector, which is then fed into a fully connected layer that projects the high-dimensional embedding into a lower-dimensional latent vector, denoted as \( \bm{L}_h \). The latent vector \( \bm{L}_h \) is then quantized and subjected to entropy encoding for more efficient compression. More details will be introduced in \ref{sec:entropy}. 

The decoding process mirrors the encoding process. Initially, we project \( \bm{L}_h \) to a higher dimension using a fully connected layer and then reshape it into \( k \) vectors, each with the same dimension as \( \bm{\tilde{e}}_i \). These reshaped vectors are processed using the same attention mechanism described in Equation \ref{eq:atten}. Finally, the output vectors are fed into the decoder \( D \) separately, yielding the reconstructed hyper-block data \( \bm{y}_{i:i+k} \).

The hyper-block autoencoder excels at high compression ratios by capturing shared information from multiple data blocks and encoding it into a single latent vector. However, this approach inevitably leads to a loss of block-specific detail. To address this, we propose a block-wise residual autoencoder, which complements the hyper-block autoencoder by capturing more detailed information within each individual block.

% Attention mechanism has also been widely adopted in many vision tasks. Comparing to the traditional convolution, attention allows the model to focus on important regions within a larger size context. In this section, we developed self-attention based autoencoder to capture the context information among multiple data blocks.

\subsection{Block-Wise Residual Autoencoder}

We employ a block-wise autoencoder to capture the residual information between the original data block \(\bm{x}_i\) and the reconstructed data block \(\bm{y}_i\) obtained from the hyper-block autoencoder. Considering that the residual values are typically very small, we apply layer normalization to re-scale the data to a more suitable range before feeding it into the block-wise autoencoder. The block-wise autoencoder consists of an encoder \(E(\cdot)\) and a decoder \(D(\cdot)\), which are architecturally similar to those used in the hyper-block autoencoder, with the exception that they operate on embeddings of different dimensions. We feed the normalized residual data into the encoder, obtaining the encoded latent representation \(L_b\), which encapsulates block-specific information. Subsequently, the decoder takes the latent representation to reconstruct the residual data, which is then added to the initial reconstruction \(\bm{y}_i\), yielding a more refined reconstruction denoted as \(\bm{x}^R_i\). The process of the residual autoencoder is formally described as follows:
\begin{equation}
    \bm{L}_b = E\left({norm}\left(\bm{x}_i-\bm{y}_i \right)\right),
\end{equation}
\begin{equation}
    \bm{x}^R_i = D(\bm{L}_b) + \bm{y}_i.
\end{equation}
\subsection{Error Bound Guarantee}
\begin{algorithm}
\caption{The GAE Algorithm}
\label{alg:GAE}
\begin{flushleft}
\textbf{Input:} Input data $ \Omega = \left\{\boldsymbol{x}_{i}\right\}_{i=1}^{N}$, reconstructed data $\Omega^R = \left\{\boldsymbol{x}_{i}^{R}\right\}_{i=1}^{N}$, target error bounds $\tau$.

\textbf{Output:} Corrected reconstruction $ \Omega^G = \left\{\boldsymbol{x}^{G}_{i}\right\}_{i=1}^{N}$, coefficients $C = \left\{\boldsymbol{c}_{i}\right\}_{i=1}^{N}$, indices $I = \left\{I_{i}\right\}_{i=1}^{N}$ where $I_i$ is an index set, basis matrix $\bm{U}$.
\end{flushleft}
\begin{algorithmic}[1]
\State Run PCA on the residual $\Omega-\Omega^R$, obtaining basis matrix $\bm{U}$
\For{$i=1$ \textbf{to} $N$}  
\State $\boldsymbol{x}\gets \boldsymbol{x}_{i}$, $\boldsymbol{x}^{R}\gets \boldsymbol{x}_{i}^{R}$.
\State Compute $\ell_{2}$ norm $\delta=\left\|\boldsymbol{x}-\boldsymbol{x}^{R}\right\|_{2}$.
\If{$\delta > \tau$}
\State Project residual $\boldsymbol{c}=\bm{U}^{T}(\boldsymbol{x}-\boldsymbol{x}^{R})$ and sort $c_{k}^{2},~\forall k$.
\State $M\gets 1$
\While{$\delta>\tau$}
\State $\boldsymbol{c}_{s},\bm{U}_{s}\gets$ Top $M$ coefficients in $\boldsymbol{c}$ and corresponding basis vectors in $\bm{U}$.
\State $\boldsymbol{c}_{q}$ $\gets$ Quantize($\boldsymbol{c}_{s}$)
\State $\boldsymbol{x}^{G}\gets \boldsymbol{x}^{R}+\bm{U}_{s}\boldsymbol{c}_{q}$.
\State $\delta \gets \left\|\boldsymbol{x}-\boldsymbol{x}^G\right\|_{2}$
\State $M\gets M+1$
\EndWhile
\State $\boldsymbol{c}_{i}\gets \boldsymbol{c}_{q}$
\State $I_{i}\gets$ Index set for $\boldsymbol{c}_{q}$
\State $\boldsymbol{x}^{G}_{i}\gets \boldsymbol{x}^{G}$
\EndIf
\EndFor
\end{algorithmic}
\end{algorithm}

We aim to minimize the reconstruction \(\ell_{2}\)-norm errors for all instances, denoted as \(\left\|\bm{x} - \bm{x}^G\right\|_{2}\), where \(\bm{x}\) represents the original data, and \(\bm{x}^G\) represents the reconstructed data \(\bm{x}^R\) from the block-wise autoencoder after applying the error bound guarantee. We denote the process of guaranteeing the error bound of the autoencoder output as GAE. Although the data is compressed block by block using autoencoders, it is not necessary to use the same block size for post-processing. Instead, we define different block formats for various data types based on the characteristics of the dataset and user requirements. The data blocks are further flattened into vectors for post-processing. For simplicity, we continue to use \(\bm{x}\), \(\bm{x}^R\), and \(\bm{x}^G\) to represent the flattened block data used for GAE.

After obtaining the reconstructed data, we apply Principal Component Analysis (PCA) to the residuals of the entire dataset to extract the principal components, or basis matrix, denoted as \(\bm{U}\). These basis vectors are sorted in descending order according to their corresponding eigenvalues. Each flattened block data is treated as a single instance, and the basis matrix is computed at the block level. To ensure the error bound for each block, the residual of each patch is projected onto the space spanned by \(\bm{U}\), and the leading coefficients are selected such that the \(\ell_{2}\) norm of the corrected residual falls below the specified threshold \(\tau\). These coefficients, representing the residual, are derived from the equation:
\begin{equation}
\boldsymbol{c} = \bm{U}^{T}(\boldsymbol{x} - \boldsymbol{x}^R).
\end{equation}
It is important to note that complete recovery of the residual \(\boldsymbol{x} - \boldsymbol{x}^R\) can be achieved by computing \(\bm{U}\boldsymbol{c}\), yielding the coefficient vector \(\boldsymbol{c} \equiv \left[c_{1}, \ldots, c_{D}\right]\). Given that the error bound criterion is based on \(\ell_{2}\), we compute \(\{c_{k}^{2}\}_{k=1}^{D}\) and sort the positive values. The coefficients are therefore sorted in the order of their contribution to the error. The top \(M\) coefficients and corresponding basis vectors are selected to satisfy the target error bound \(\tau\). To minimize the storage cost of these coefficients, we compress the selected coefficients \(\boldsymbol{c}_{s}\) using quantization followed by entropy coding, which will be introduced in the next section. These coefficients are quantized before being used for reconstructing the residual. The corrected reconstruction \(\boldsymbol{x}^{G}\) is given by:
\begin{equation}
\boldsymbol{x}^{G} = \boldsymbol{x}^{R} + \bm{U}_{s}\boldsymbol{c}_{q},
\end{equation}
where \(\boldsymbol{c}_{q}\) represents the set of selected coefficients \(\boldsymbol{c}_{s}\) after quantization, and \(\bm{U}_{s}\) denotes the set of selected basis vectors. We increase the number of coefficients until we achieve \(\left\|\boldsymbol{x} - \boldsymbol{x}^{G}\right\|_{2} \leq \tau\). The GAE algorithm is shown step by step in Algorithm \ref{alg:GAE}.
% Therefore, to guarantee the error bound, we need to store the basis matrix, the selected coefficients \(\boldsymbol{c}_{q}\) for each patch of data, and their basis vector indices.

\subsection{Efficient Coefficients Storage}
\label{sec:entropy}

To efficiently store latent space data, we employ a compression technique involving floating-point quantization followed by entropy encoding. First, we uniformly quantize the latent coefficients into discrete bins. This discretization process represents all values within a bin by its central value, transforming the originally continuous data into a discrete form. Subsequently, we apply Huffman coding to compress these quantized coefficients. Huffman coding assigns shorter codes to more frequently occurring quantized coefficients, optimizing data representation and achieving higher compression efficiency. We apply this floating-point data compression separately to the latent space of the hyper-block autoencoder and the block-wise autoencoder.

\begin{figure}[h!]
    \centering
    \includegraphics[width=0.45\textwidth]{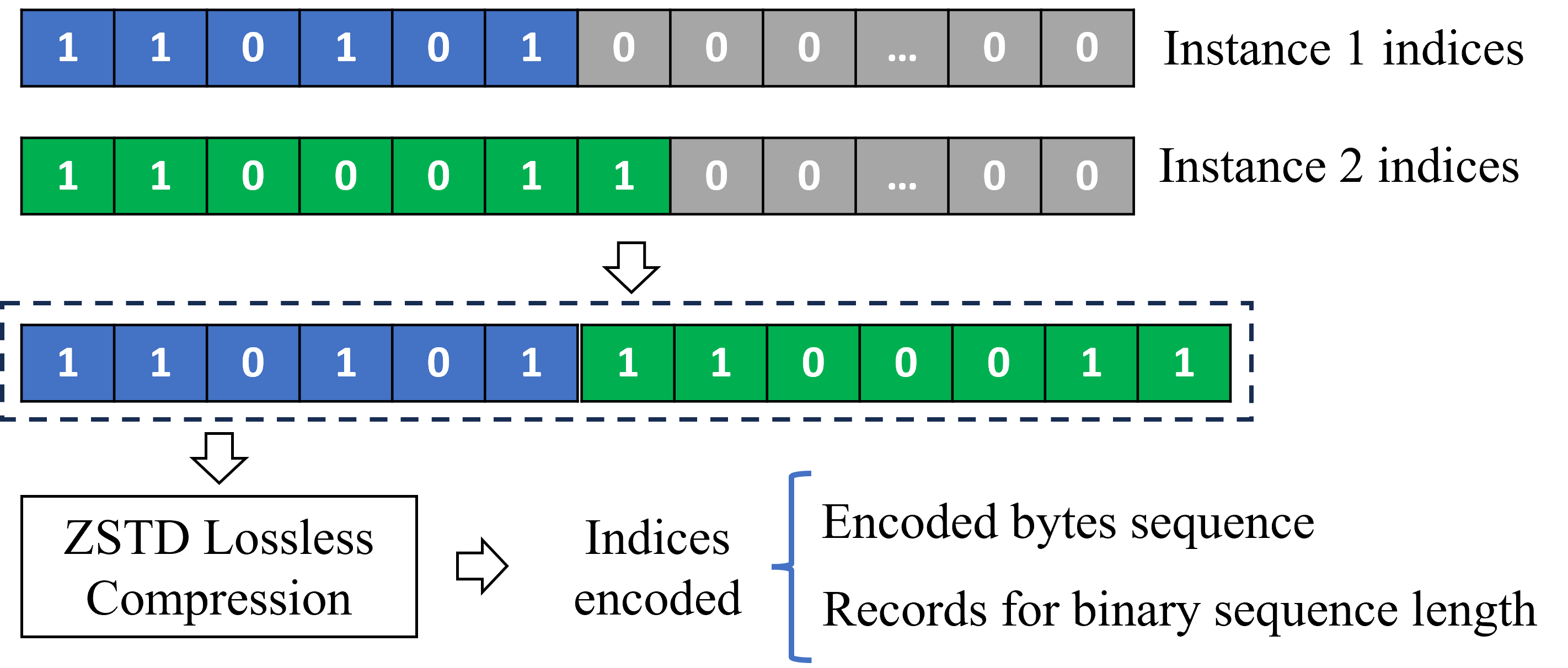}  % Replace with the actual path to your image
    \caption{Indices Encoding}
    \label{fig:indices}
        \vspace*{-0.3cm}
\end{figure}

To improve the compression ratio while ensuring an error-bound guarantee, we apply the same technique to compress the selected coefficients. In addition to saving the PCA coefficient data, we must also store the indices of the principal vectors corresponding to these coefficients. However, directly applying entropy coding to these integer indices yields little improvement. Instead, we represent the index selection as a binary sequence as shown in Figure \ref{fig:indices}, where `1' indicates the corresponding vector is selected, and `0' indicates it is not selected. Instead of storing the full sequence, we save only the shortest prefix containing all the `1's, along with an integer that records the length of this prefix. These binary sequences are further concatenated and compressed by lossesless compressor ZSTD \cite{collet2018zstandard}.  

\section{Experimental Results}\label{sec:experiment}
This section presents the experimental results of applying our compression pipeline to the datasets obtained from the S3D, E3SM, and XGC applications. We provide a thorough explanation of the evaluation metrics and outline the key characteristics of the dataset, along with detailing the experimental setup. We establish a baseline method for comparison with our approach. Subsequently, we evaluate the accuracy and effectiveness of our compression method and compare it with existing methods in the literature.

\subsection{Evaluation Metrics and Datasets}

We employ the Normalized Root Mean Square Error (NRMSE) as a relative error criterion to evaluate the quality of reconstruction, taking into account that different datasets may span various data ranges. The NRMSE is defined in Equation \ref{eq:nrmse}:
\begin{equation}
\label{eq:nrmse}
\mathrm{NRMSE}\left(\Omega,\Omega^{G}\right)=\frac{\sqrt{\left\|\Omega-\Omega^{G}\right\|_{2}^{2}/N_d}}{\max\left(\Omega\right)-\min\left(\Omega\right)},
\end{equation}
where $N_d$ is the number of data points in the dataset, and $\Omega$ and $\Omega^{G}$ represent the entire original dataset and the reconstructed datasets with GAE post-processing, respectively.

% Additionally, we evaluate data reduction through the compression ratio, defined as the ratio between the sizes of the primary data (PD) and the compressed output. The compressed output includes the encoded representation of AE encoders, the decoder size of the hyperblock AE and block-wise AE, encoded coefficients with their corresponding basis indicators for the error-bound guarantee post-processing, and all the dictionaries for entropy coding.

% We do not factor in the size of the AE, as it consists of fixed and one-time dictionaries, which become negligible with increasing data size.
\subsection{Compression Ratio}

The overall target for scientific data compression can be summarized as achieving a higher compression ratio while adhering to a user-defined error bound. Given a multi-dimensional scientific dataset $ \Omega $, a lossy compressor $C_e$, and a corresponding decompressor $D$, the compressed data can be generated by $L = C( \Omega )$. The reconstructed data can then be obtained by $ \Omega^G = D(L)$. The compression ratio can be expressed as:
\begin{equation}
    \text{Compression Ratio} = \frac{\text{Size}( \Omega )}{\text{Size}(L)}.
\end{equation}
\paragraph*{\textbf{S3D} Dataset}
% Our experiments use a dataset spanning five years at a temporal resolution of 6 hours and stored in \texttt{float32} precision.

We now briefly introduce the S3D dataset, which represents the compression ignition of large hydrocarbon fuels under conditions relevant to homogeneous charge compression ignition (HCCI), as detailed in \cite{Yoo11}. The dataset comprises a two-dimensional space of size $640\times640$, collecting data over 50 time steps uniformly from $t = 1.5$ to $2.0$ ms. A 58-species reduced chemical mechanism \cite{Yoo11} is used to predict the ignition of a fuel-lean $n$-heptane/air mixture. Thus, each tensor corresponds to 58 species, resulting in a 4D dataset with the shape $58 \times 50\times 640\times 640$. The research conducted by \cite{jung2024application} has demonstrated a strong correlation among the 58 species under examination. In line with their methodology, we compressed 58 species together and blocked the dataset into 4D tensors with a shape of $58\times 5\times 4\times 4$, where we combined 5 timesteps of data with a spatial size of $4\times 4$. All 58 species were aggregated into 4D tensors. Since the ranges of the species vary significantly, each species was normalized to have a mean of 0 and a range of 1. Continuous, non-overlapping blocks of 10 along the temporal dimension were aggregated to construct hyperblocks which were subsequently fed into the proposed autoencoders. Subsequently, we guarantee the error bound of each species separately with block size $5\times 4\times 4$ for each species. We calculate the NRMSE of each species and compression ratio in the original domain of 58 species. We report the mean value of NRMSE of the 58 species for the S3D dataset.

% To compare with the baseline SZ, we run SZ on each of the species separately using the whole spatiotemporal block with a shape of $50\times 640\times 640$, obtaining the compression ratio and corresponding NRMSE for each species.

\paragraph*{\textbf{E3SM Dataset}}
The E3SM (Energy Exascale Earth System Model) serves as a cutting-edge computational framework meticulously crafted to simulate Earth's climate system with unparalleled fidelity and resolution. Our method's efficacy is evaluated utilizing climate data generated by the E3SM simulator. The dataset originates from a high-resolution (HR) configuration atmosphere E3SM simulation. Employing a grid resolution of 25 km, each variable yields approximately 350,000 float32 data points per hour. The data simulation adopts a grid spacing of $0.25^{\circ}$. Through Cube-to-Sphere Projections, we project Earth's coordinate-based data onto a planar surface, yielding an image dataset with dimensions of $720 \times 240 \times 1440$. Here, 720 denotes the number of timesteps/hours, with each timestep possessing a spatial resolution of $240 \times 1440$. 
For evaluation, we utilize the sea-level pressure (PSL) climate variable. Initially, we normalize the PSL dataset using z-scores and subsequently partition it into spatiotemporal blocks of $6\times16\times16$, where six timesteps are combined with a spatial resolution of $16 \times 16$. We further aggregate continuous 5 blocks along temporal dimension into one hyperblock. we utilize GAE post-processing with block size $16 \times 16$ on the reconstructed data to guarantee the error bound.

% To benchmark against SZ, we input the entire spatiotemporal dataset, sized $720 \times 240 \times 1440$, into the SZ compressor. 

\paragraph*{\textbf{XGC} Dataset}

The X-Point Inclusion Gyrokinetic (XGC) simulation, introduced in~\cite{chang2008spontaneous,ku2009full}, serves as a powerful tool for modeling magnetically confined thermonuclear fusion (tokamak) plasmas. It addresses gyrokinetic particle-in-cell (PIC) equations across five dimensions, encompassing three spatial dimensions and two velocity dimensions. Within XGC, we utilize a particle distribution function referred to as $F$-data.

The D3D \( F \)-data consists of 8 toroidal cross-sections, each with 16,395 mesh nodes per cross-section at each simulation timestep. Every mesh node contains a \( 39 \times 39 \) 2D velocity histogram of particle counts, resulting in a dataset size of \( 8 \times 16,395 \times 39 \times 39 \). For the training process, the dataset is normalized by z-scores across the entire dataset. Since the data across the 8 toroidal cross-sections are highly correlated, we aggregate 8 histograms into a single hyperblock at the same location on each cross-section. For GAE post-processing, we consider each $39 \times 39$ histogram as a block.

% For comparison with SZ, we compress the entire 4-dimensional XGC dataset using SZ and calculate the NRMSE and compression ratio over the entire dataset.

\begin{table}[h!]
\centering
\begin{tabular}{c c c c c}
\hline
Application & Domain & Dimensions & Total Size \\ \hline
S3D & Combustion & $58 \times 50 \times 640 \times 640$ & 9.5 GB\\ 
E3SM & Climate &  $720 \times 240 \times 1440$ & 1.0 GB\\ 
XGC & Plasma &  $8 \times 16395 \times 39 \times 39$ & 1.6 GB\\ \hline
\end{tabular}
\caption{Datasets Information}
\label{tab:dataset}
\vspace*{-0.5cm}
\end{table}

\subsection{Experimental Setups}

Our experiments were conducted on the Hypergator Supercomputer. We utilized 10 CPUs and 2 NVIDIA A100 GPUs for model training and testing. The entire framework was implemented with the PyTorch library.

The hyper-block autoencoder (HBAE) was trained first, followed by the block-wise autoencoder (BAE). A Mean Square Error (MSE) loss function was employed to minimize the error between the original data and the reconstructed data. We used the Adam optimizer with a learning rate of 0.001 to update the model parameters during training. The latent dimension of the hyper-block autoencoder was set to 128 for the S3D dataset and 64 for the XGC and E3SM datasets. The latent dimension of the block-wise autoencoder was set to 16 for all three datasets. Different error bounds were set during the GAE post-processing to obtain data points with varying compression ratios. To calculate the compression ratio, we considered the latent spaces of both the hyper-block autoencoder and the block-wise autoencoder, as well as the PCA coefficients and corresponding index information used in GAE post-processing.

\subsection{Ablation Study}
To explore the performance of the proposed method, we conduct several ablation study on the S3D dataset. We use an block-based compressor as the baseline which divides the original data into blocks and compress the block data with a set of cascaded fully connected layers. In this section, we didn't apply error bound guarantee to the reconstructed data or quantization on latent latent. we obtain different compression ratio by varying the dimension of latent space.

\begin{figure}[h!]
    \centering
    \includegraphics[width=0.4\textwidth]{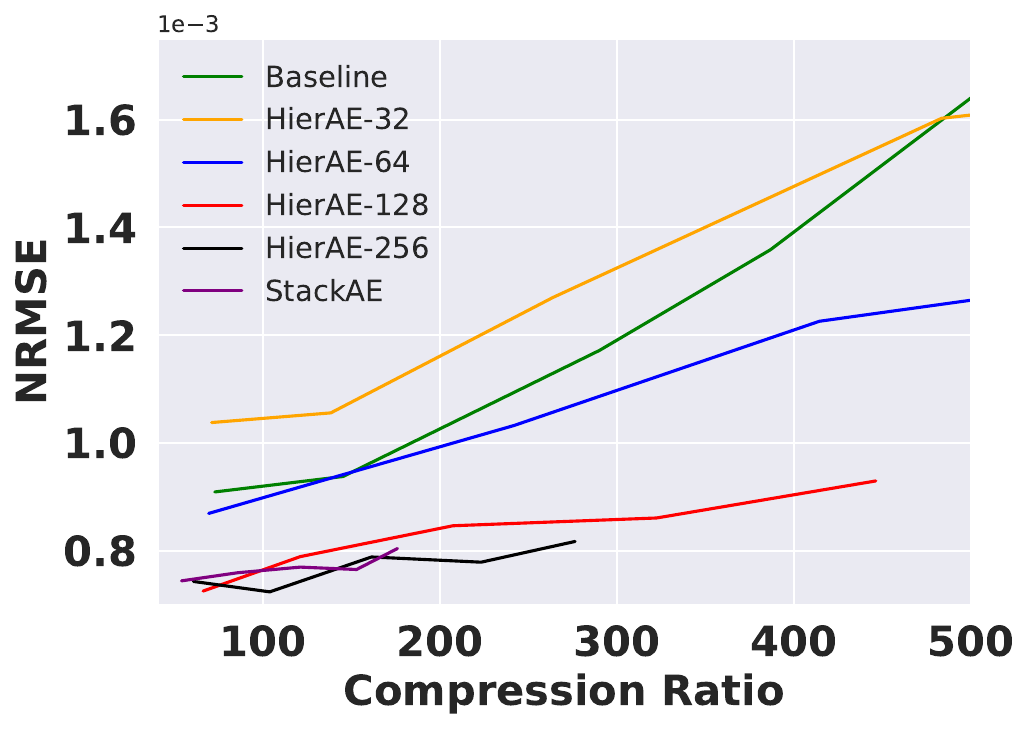}  % Replace with the actual path to your image
    \caption{Ablation study of latent size on S3D dataset}
    \label{fig:ablation_latent}
    \vspace*{-0.5cm}
\end{figure}

To investigate how the HBAE latent dimension affects compression performance, we first vary the latent size of the BAE from 8 to 128. For the HBAE, we compare latent dimensions of 32, 64, 128, and 256 to generate different performance curves, denoted as `HierAE-N'. Additionally, we include the performance of the baseline approach for comparison, denoted as `Baseline'. As shown in Figure \ref{fig:ablation_latent}, compression performance improves clearly with the increasing size of the hyper-block latent dimension. However, we also observe that larger latent sizes lead to longer training times and less stability during training. We also tested adding more residual block-wise autoencoders in a stacked manner. The result using one hyperblock autoencoder cascaded with two residual block-wise autoencoders are shown in Figure \ref{fig:ablation_latent}, denoted as `StackAE'. The result indicates that stacking additional block-wise autoencoders on the residuals contributes little to the compression performance. Therefore, we use the configuration with one hyperblock autoencoder and one residual block-wise autoencoder as the final setup.

\begin{figure}[h!]
    \centering
    \includegraphics[width=0.4\textwidth]{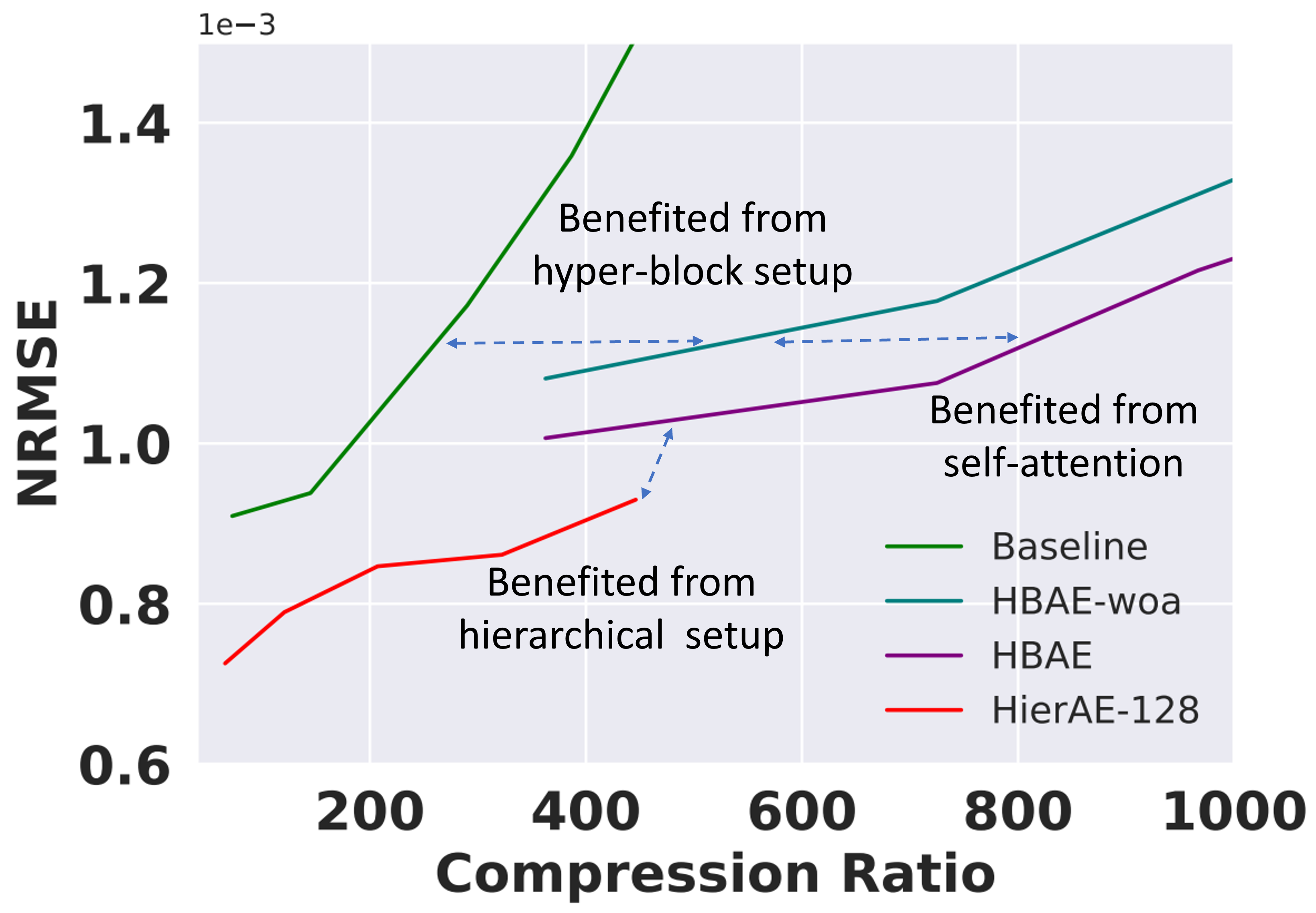}  % Replace with the actual path to your image
    \caption{Ablation study of each model component}
    \label{fig:atten}
    \vspace*{-0.6cm}
\end{figure}

\begin{figure*}[h!]
    \centering
    \begin{subfigure}[b]{0.32\textwidth}
        \centering
        \includegraphics[width=\textwidth]{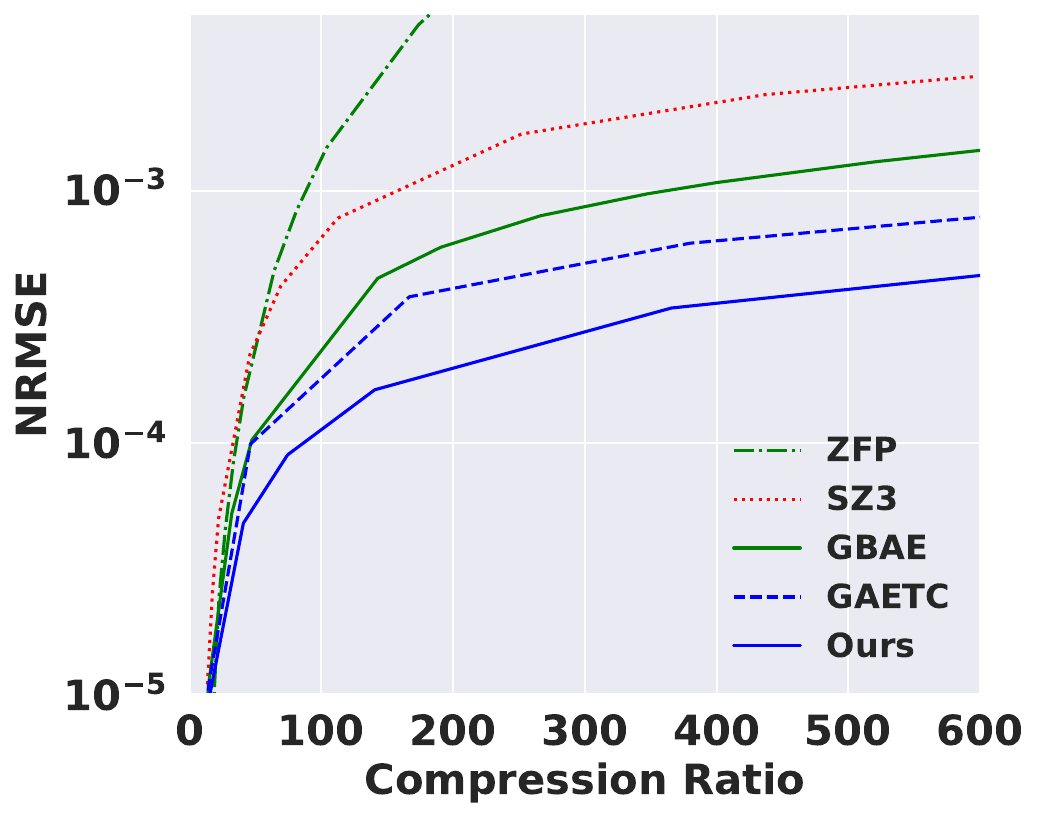}
        \caption{Results on S3D dataset}
        \label{fig:sz}
    \end{subfigure}
    \hfill
    \begin{subfigure}[b]{0.32\textwidth}
        \centering
        \includegraphics[width=\textwidth]{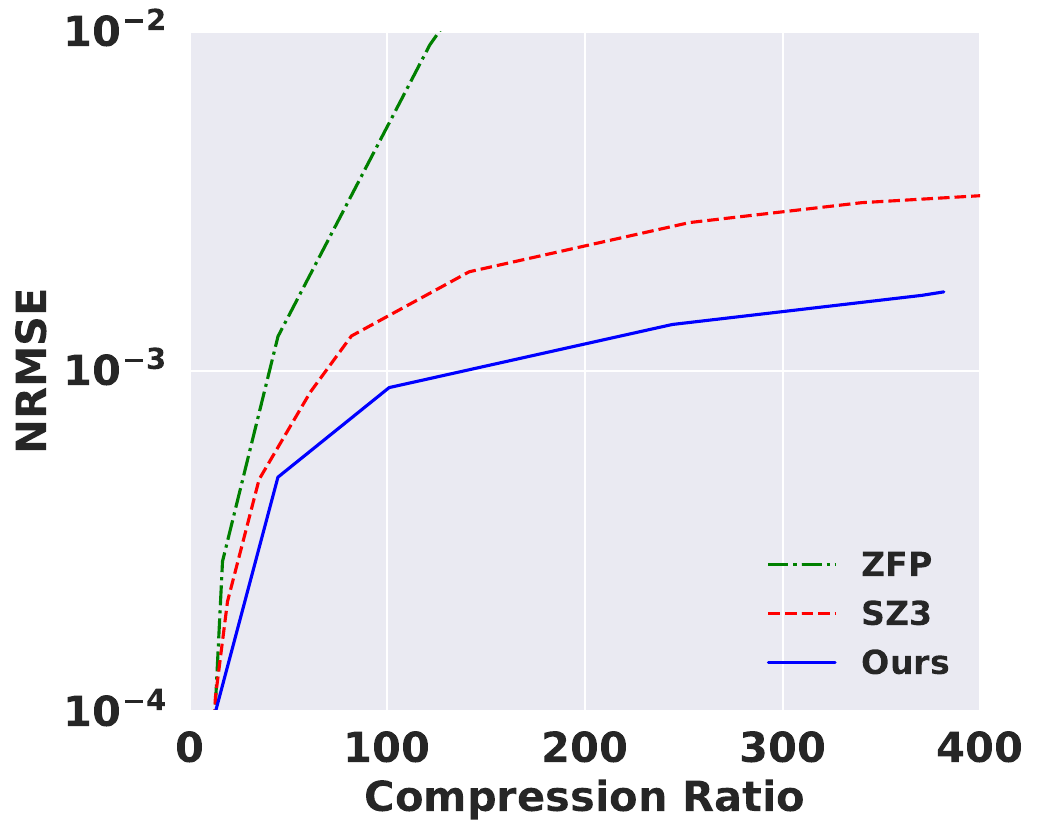}
        \caption{Results on E3SM dataset}
        \label{fig:mgard}
    \end{subfigure}
    \hfill
    \begin{subfigure}[b]{0.32\textwidth}
        \centering
        \includegraphics[width=\textwidth]{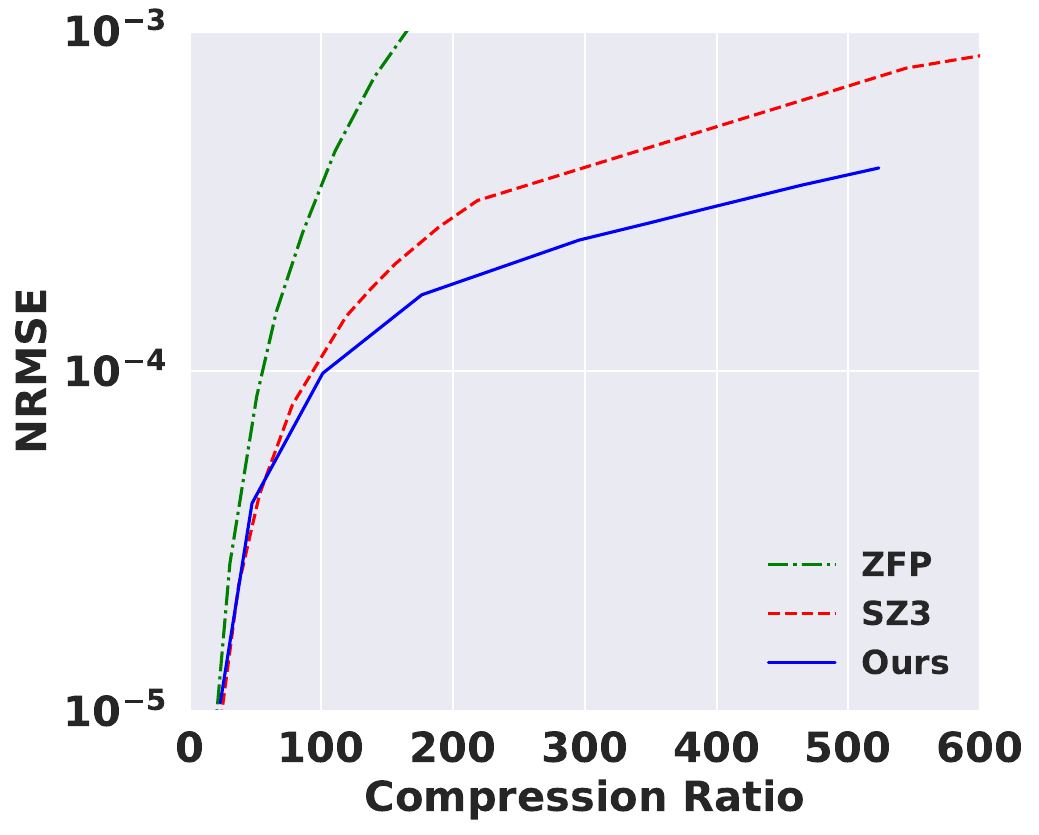}
        \caption{Results on XGC dataset}
        \label{fig:zfp}
    \end{subfigure}
    \caption{Comparisons for different methods on S3D, E3SM and XGC dataset.}
    \label{fig:main_curve}
    \vspace*{-0.3cm}
\end{figure*}

\begin{figure}[h!]
    \vspace*{-0.4cm}
    \centering
    \includegraphics[width=0.4\textwidth]{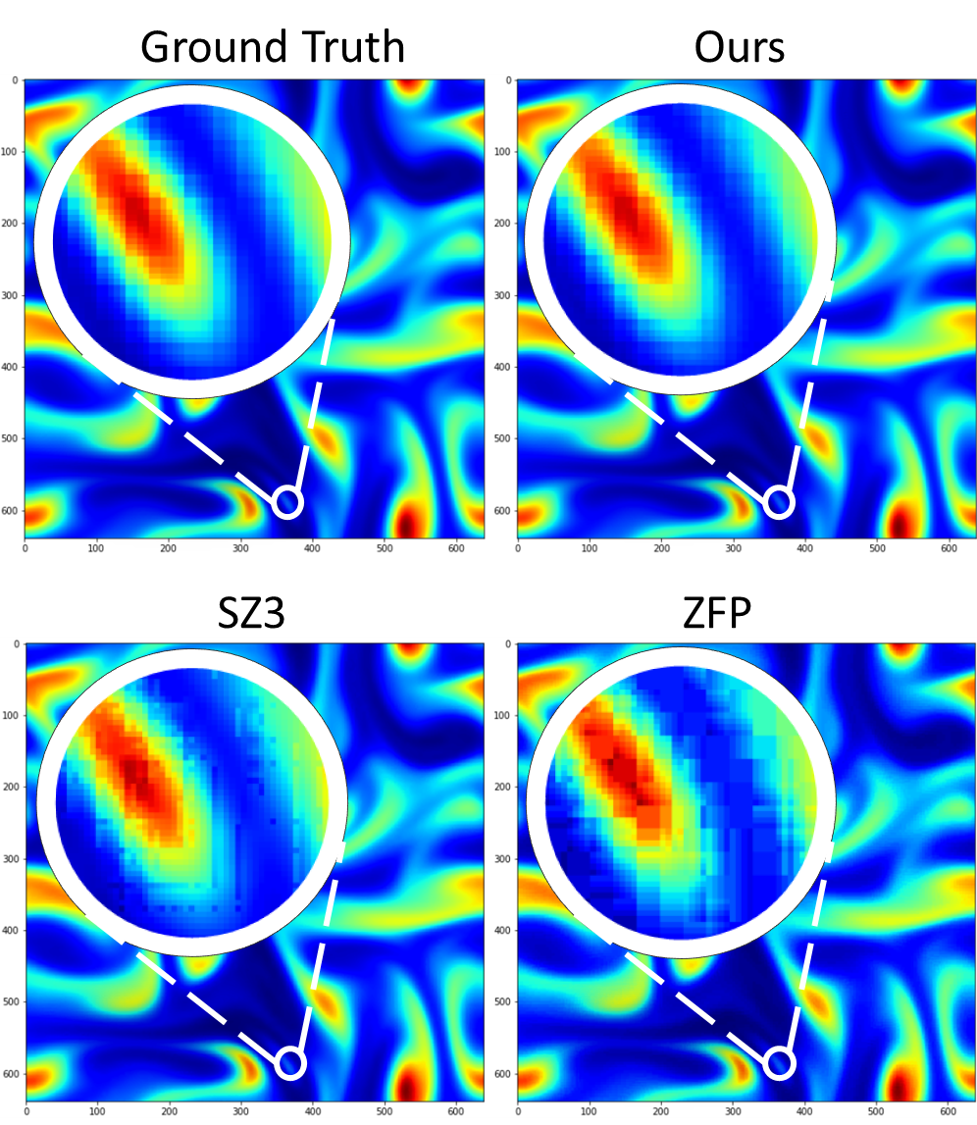}  % Replace with the actual path to your image
    \caption{Visualization of S3D data reconstructed using the proposed method, SZ and ZFP at the compression of 100.}
    \label{fig:vis}
    \vspace*{-0.5cm}
\end{figure}

% Therefore, we recommend 128-dimensional latent space for hyperblock autoencoder.

We further explore the impact of the hierarchical setup and the self-attention mechanism. First, we evaluate the reconstruction error of the HBAE without the residual BAE, denoted as `HBAE' in Figure \ref{fig:atten}. Next, we remove the self-attention module from the HBAE and train another version without self-attention, denoted as `HBAE-woa'. For comparison, we also include the performance of the baseline approach and the proposed hierarchical method in Figure \ref{fig:atten}. Comparing the baseline method with the hyper-block AE without self-attention, the hyper-block setup clearly improves the compression ratio at the same compression error. The self-attention mechanism further enhances the performance of the HBAE. Additionally, by adding the residual BAE, the NRMSE is reduced to a lower level compared to `HBAE'. These experimental results demonstrate the effectiveness of each component in the proposed hierarchical method.

\subsection{Results}

To efficiently compress the latent space of autoencoders, we tested different quantization bin sizes across three datasets. Larger quantization bins led to better compression ratios but also resulted in higher reconstruction errors. To evaluate the impact of latent space quantization on both the HBAE and the BAE, we quantized the latent space of one autoencoder while leaving the latent space of the other unquantized. We then calculated the reconstruction error from the output of the residual BAE. Table \ref{table:quantization_errors} presents the results of various quantization setups on these datasets. For example, the row labeled "HBAE" indicates that the latent space of the HBAE was quantized while the latent space of the BAE was not. The table showed that the HBAE was more sensitive to latent space quantization with respect to the reconstruction error. As the quantization bin size increased, the reconstruction error grew more rapidly compared to the BAE. To balance the compression ratio and reconstruction error, we selected a quantization bin size of 0.005 for both autoencoders on S3D, 0.1 for both autoencoders on XGC, and 0.01 for the HBAE and 0.1 for the BAE on E3SM.

\begin{table}[h]
\centering
\begin{tabular}{|c|c|c|c|c|c|c|c|}
\hline
\multirow{3}{*}{\textbf{S3D}} &  \textbf{Bin Size} & \textbf{0.005} & \textbf{0.01} & \textbf{0.05} & \textbf{0.1} & \textbf{0.5} \\ \cline{2-7}
& {HBAE}  & 8.6e-4 & 8.8e-4 & 9.2e-4 & 1.9e-3 & 3.5e-3 \\ \cline{2-7}
& {BAE}  & 8.6e-4 & 8.7e-4 & 8.8e-4 & 1.2e-3 & 2.0e-3 \\ 
\hline\hline

\multirow{3}{*}{\textbf{E3SM}} &  \textbf{Bin Size} & \textbf{0.001} & \textbf{0.005} & \textbf{0.01} & \textbf{0.05} & \textbf{0.1} \\ \cline{2-7}
& {HBAE} & 1.6e-3 & 1.6e-3 & 2.3e-3 & 3.6e-3 & 1.3e-2 \\ \cline{2-7}
& {BAE} & 1.6e-3 & 1.6e-3 & 1.6e-3 & 1.7e-3 & 2.5e-3 \\ 
\hline\hline

\multirow{3}{*}{\textbf{XGC}} & \textbf{Bin Size} & \textbf{0.05} & \textbf{0.1} & \textbf{0.2} & \textbf{0.4} & \textbf{0.8} \\ \cline{2-7}
& {HBAE} & 8.2e-4 & 8.3e-4 & 9.0e-4 & 1.4e-3 & 2.3e-3 \\ \cline{2-7}
& {BAE} & 8.2e-4 & 8.2e-4 & 8.2e-4 & 8.3e-4 & 8.7e-4 \\ 
\hline
\end{tabular}
\caption{Reconstruction error across different datasets and autoencoders with varying bin sizes.}
\label{table:quantization_errors}
\vspace*{-0.3cm}
\end{table}
To demonstrate the effectiveness of the proposed method, we conduct several comparisons with the state-of-the-art lossy compressors on S3D, E3SM and XGC dataset. SZ3 \cite{SZ3} is a state-of-the-art scientific data compressor which works by predicting the value of each data point based on its neighbors, using a combination of linear regression and other predictive models. ZFP  \cite{zfp} is a transform-based compressor that employs a block-wise transform approach to compress floating-point data. We also compare our results with a block-based method GBAE \cite{lee2024machine} on the S3D dataset. GAETC \cite{lee2024machine} boosts the compression ratio from GBAE by adding extra tensor correction network to capture the correlation among 58 species of S3D data.

As shown in Figure \ref{fig:main_curve}, our method outperforms or achieves comparable performance to existing methods. Specifically, on the S3D dataset, our method achieves more than a 2-fold compression ratio at an NRMSE of \(1 \times 10^{-4}\) compared to SZ3, and achieves a 2-8 fold increase in compression ratio within the NRMSE range of \(1 e^{-3}\) to \(1 e^{-4}\). Our method excels at efficiently capturing the correlation among multiple species in the S3D dataset. Compared to BAE and GAETC, which are designed to capture high-dimensional correlations, our model shows significant improvement on the S3D dataset. Our method succeeds in effectively capturing inter-block correlations. Additionally, on the E3SM dataset using single pressure variable, the proposed method demonstrates a higher compression ratio compared to SZ3 and ZFP, achieving up to a 3-fold improvement with the same NRMSE as SZ3. For the XGC dataset, our method achieves up to 2-fold improvement in compression ratio at high compression ratio level and delivers comparable performance at lower compression ratio level when compared to SZ3. The evaluation results highlight the efficiency of our method in compressing high-dimensional data and capturing both inter-block and inter-dimensional correlations.

\subsection{Visualization of Reconstructed Data}

\begin{figure}[h!]
  \vspace*{-0.2cm}
    \centering
    \includegraphics[width=0.35\textwidth]
    {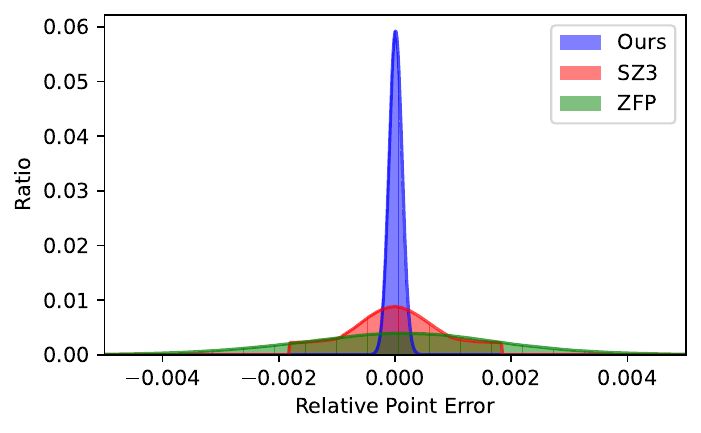}  % Replace with the actual path to your image
    \vspace*{-0.2cm}
    \caption{Histogram of relative point error for 3 compressors.}
    \label{fig:distribution}
    \vspace*{-0.5cm}
\end{figure}

We further compare the compression quality of multiple compressors by visualizing reconstructed images alongside the original data. We use the first species in S3D dataset as an example, and show reconstructed images from different compressors. For a fair comparison, each compressor processes the data with a fixed compression ratio of around 100. The NRMSE errors for our method, SZ3, and ZFP are $1.2 e^{-4}$, $8.2 e^{-4}$, and $1.8e^{-3}$, respectively. Our method achieves the lowest NRMSE error compared to the other compressors. As shown in Figure \ref{fig:vis}, we zoom in on a critical region for comparison. The proposed method provides the highest image quality and the lowest distortion relative to the original data. To visualize the reconstruction error for each data point, we calculated the residuals from three compressors and normalized them by dividing by the range of the original data, referring to this as the relative point error. Figure~\ref{fig:distribution} presents the histogram of relative point errors from all three compressors at a compression ratio of approximately 100. Compared to the other compressors, our method shows relative point errors predominantly concentrated at lower values.

\subsection{Visualization at each element of the tensor}

\begin{figure*}[!htb]
    \centering
    \includegraphics[width=\textwidth]{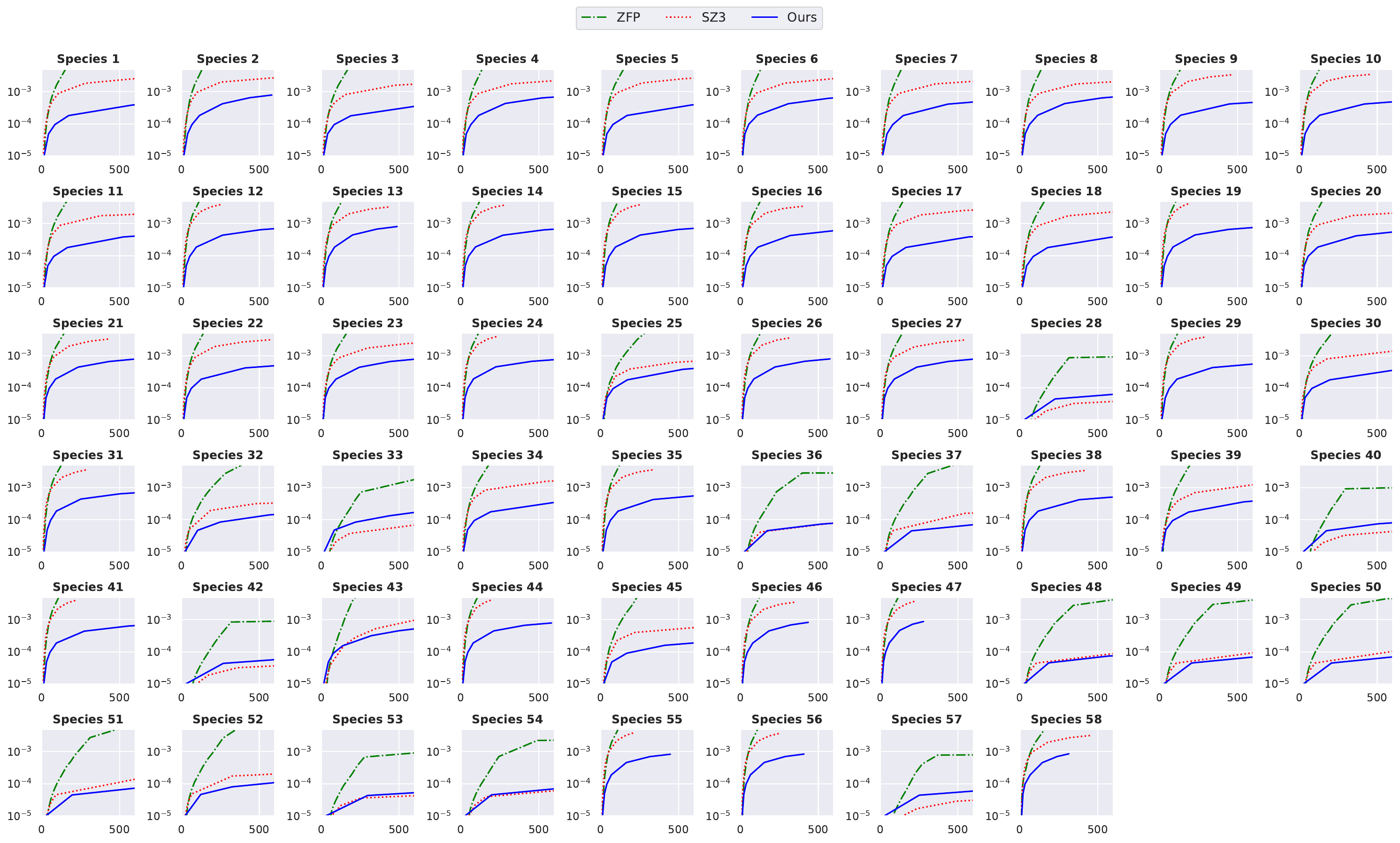}
    \caption{Reconstruction Error for Each Species on the S3D Dataset.}
        \vspace*{-0.6cm}
    \label{fig:s3d_each}
\end{figure*}

So far, we have presented the mean NRMSE for 58 species from the S3D dataset. To further illustrate the compression performance for each species, Figure \ref{fig:s3d_each} shows the reconstruction NRMSE error versus compression ratio curves for ZFP, SZ3, and our method.  Our method outperforms SZ3 and ZFP for most species, achieving better overall compression performance across all species. To calculate the compression ratio for each species in the S3D dataset, we assume that the latent space cost of the autoencoders is amortized equally across all species. We then use the amortized latent cost, along with the cost of ensuring error bound guarantees, to compute the compression ratio.

% Ensure figure stays on the same page

\section{Related Work}\label{sec:relwork}
% The literature on image and video compression techniques is large: please see survey articles: \cite{saha2000image} for image compression; \cite{bhaskaran1997image} for video compression; and \cite{ma2019image} for more recent neural network-based compression. We mainly focus on reduction and compression approaches for scientific applications because techniques for image and video compression are not directly applicable as discussed earlier.
%We adopt a bifurcated approach to discussions of previous work. 
%\sanjay{A sentence about why image processing literature is not relevant}

Error-bounded lossy compression is widely regarded as the most effective technique due to its ability to provide reliability, which is particularly valuable for scientific applications. This technique ensures that compression maintains error control within specified bounds. Lossy compression methods can be categorized into several types, including transform-based methods, prediction-based methods, machine learning-based methods and hybrid methods.

\paragraph*{Transform-Based Methods}
ZFP \cite{fox2020stability} is a transform based compression model that splits a dataset into a set of unoverlapped 4D blocks. Each block is decorrelated using a nearly orthogonal transform.  TTHRESH \cite{ballester2019tthresh} is a dimensionality reduction-based model and uses higher order singular value decomposition (HOSVD) to reduce the dimension of the data according to importance. MGARD \cite{MGARD_2, gong2023mgard} offers error-controlled lossy compression through a multigrid approach. It operates by transforming floating-point scientific data into a set of multilevel coefficients, facilitating efficient compression while maintaining error control. 
% The multigrid technique enables effective representation of the data across multiple resolution levels, enhancing compression performance while preserving essential information for scientific analysis and interpretation. 
In \cite{Archibald2023}, the weak SINDy algorithm \cite{messenger2021weak2} that estimates and identifies an underlying dynamic system and an orthogonal decomposition are combined to compress streaming scientific data. 

\paragraph*{Prediction-Based Methods}
DPCM \cite{mun2012dpcm} is a widely used prediction-based lossy compression technique that predicts each sample based on the previous sample and encodes the difference between the predicted and actual values. Interpolative Predictive Coding is commonly employed in video compression standards such as MPEG \cite{boyce2021mpeg} , where each sample is predicted based on neighboring samples, and the prediction error is encoded. Recently, SZ \cite{SZ_3} stands out for its superior performance among prediction-based compression methods. SZ predicts each point based on its adjacent data points. Various versions of SZ have been developed using different methods to enhance compression quality. FAZ \cite{liu2023faz} is a comprehensive compression framework that has functional modules and leverages prediction models and wavelets. 

\paragraph*{Machine Learning-Based Methods}

In recent years, machine learning-based compression methods have demonstrated superior performance compared to traditional approaches, particularly in terms of the fidelity of reconstruction. Techniques such as, variational autoencoders (VAE) \cite{kingma2013auto}, along with their variations \cite{minnen2018joint,kingma2021variational}, have played pivotal roles in achieving remarkable compression results. \cite{minnen2018joint} combines autoregressive and hierarchical priors, optimizing an autoregressive component that predicts latent variables based on their causal context, alongside a hyperprior and the underlying autoencoder. QuadConv \cite{QuadConv} is developed based on convolutional autoencoders that perform convolution via quadrature for non-uniform and mesh-based data. 

\paragraph*{Hybrid Methods}

AE-SZ \cite{liu2021exploring} integrates autoencoders into the SZ compression framework, utilizing a blockwise architecture to divide data into compact fixed-size blocks for both offline training and online compression. It dynamically switches between SZ and autoencoder based on their performance, selecting the method that yields better results.\cite{li2024hybrid} combines a machine learning method with MGARD to achieve differential compression for different regions, thereby improving the compression ratio while also ensuring the preservation of quantity-of-interests. SRN-SZ \cite{liu2023srn} stands out for its integration of machine learning-based super-resolution with SZ, which ensures error control by employing the hierarchical data grid expansion concept.  

\paragraph*{Self-Attention}
Self-attention mechanism was originally proposed by ~\cite{vaswani2017attention} to address the challenges of capturing long-range dependencies and contextual relationships in sequence data. Self-attention can be regarded as a soft weighted function that dynamically adjusts the importance of different elements in a sequence based on their relationships. Recent works, such as \cite{zhou2023srformer} and \cite{liu2021swin}, have introduced self-attention mechanisms to tackle computer vision problems, demonstrating their effectiveness in capturing complex patterns and relationships within visual data. Despite the success of self-attention in various domains, there has been limited focus on attention-based autoencoders specifically for data compression. This area remains relatively underexplored, particularly in the context of scientific and high-dimensional datasets.

\section{Conclusions}\label{sec:conclusion}

In this paper, we proposed a hierarchical machine learning technique for scientific data reduction with guaranteed precision. We demonstrate promising results across various scientific datasets, including Combustion, Climate, and  Fusion. The proposed method consist of the coarse-to-fine tensor blocks architecture with self-attention, the block-wise residual autoencoder, and the error bound guarantee using PCA. Our method outperforms or matches the performance of state-of-the-art lossy compressors like SZ3 and ZFP. The visualization of reconstructed data further showcases the high-quality reconstruction achieved by the proposed approach. The ablation study highlights the importance of each component in the hierarchical setup, with the self-attention mechanism and the block-wise residual autoencoder contributing to the improved compression performance. Overall, the proposed hierarchical machine learning technique demonstrates the potential for reliable and efficient data compression in scientific applications, providing a promising solution for managing the growing volume and complexity of scientific data.

\bibliographystyle{IEEEtran}
\bibliography{ref,ref2,chen,references2,klaskybib}

% Reduce spacing in the bibliography
% \setlength{\bibsep}{0pt plus 0.3ex}
%\onecolumn
\end{document}